%% file: main_5_pages.tex
\documentclass[conference, 10pt]{IEEEtran}
\IEEEoverridecommandlockouts
\usepackage{latexsym}
\usepackage{graphicx}
\usepackage{amsfonts,amssymb,amsmath}
\usepackage{textcomp}
\usepackage{gensymb}
\usepackage{siunitx}
\usepackage{xcolor}
\usepackage{tikz}
\usetikzlibrary{chains,arrows,calc,positioning}
\usepackage{etoolbox}
\usepackage{multirow}
\usepackage[caption=false, font=footnotesize]{subfig}
\input{commands.tex}

\input{acronyms.tex}
\newtoggle{conf}
\toggletrue{conf}

\def\BibTeX{{\rm B\kern-.05em{\sc i\kern-.025em b}\kern-.08em T\kern-.1667em\lower.7ex\hbox{E}\kern-.125emX}}

\begin{document}

\title{Wireless Channel Prediction in \\ Partially Observed Environments}

\author{
\IEEEauthorblockN{Mingsheng Yin, Yaqi Hu, Tommy Azzino, Seongjoon Kang, Marco Mezzavilla, Sundeep Rangan}

\IEEEauthorblockA{NYU Tandon School of Engineering, Brooklyn, NY, USA 
\\ e-mail: \{my1778, yh2829, ta1731, sk8053, mezzavilla, srangan\}@nyu.edu}

\thanks{The authors were supported 
by NSF grants 
1952180, 1925079, 1564142, 1547332, the SRC, OPPO,
and the industrial affiliates of NYU WIRELESS. The work was also supported by Remcom that provided the Wireless Insite  
software.}
}

\maketitle
\begin{abstract}
Site-specific radio frequency (RF) propagation prediction increasingly relies on models built from visual data such as cameras and LIDAR sensors.
When operating in dynamic settings, the environment may only be partially observed. This paper introduces a method to extract statistical channel models, given partial observations of the surrounding environment. We propose a simple heuristic algorithm that performs ray tracing on the partial environment and then
uses machine-learning trained predictors to estimate the channel and its uncertainty from features extracted from the partial ray tracing results. It is shown that 
the proposed method can interpolate between fully statistical models when no partial information is available and fully deterministic models when the environment is completely observed.
The method can also capture the degree of uncertainty
of the propagation predictions depending on the amount
of region that has been explored.
The methodology is demonstrated in a robotic navigation 
application simulated on a set of indoor maps
with detailed models constructed using state-of-the-art 
navigation, simultaneous localization and mapping (SLAM), and computer vision methods. 
\end{abstract}


\IEEEpeerreviewmaketitle

\section{Introduction}
There has been growing interest in combining
radio frequency (RF) sensing, such as RADAR, 
with other sensing modalities such as camera or LIDAR
\cite{park2019radar,feng2020deep}. 
Camera data have also been proposed to guide communications, such as beamforming at millimeter-wave (mmWave) frequencies \cite{xiang2019computer,alrabeiah2020millimeter}. A basic problem in these
applications is to predict RF propagation
from visual information from cameras or other
sources.

One natural approach to this problem is 
to first build a 3D model of the environment from the camera
data.  
Numerous RGBD and cloud point cameras, along with
different view synthesis pipeline algorithms, are now widely commercially available for this purpose \cite{li2021igibson, Matterport3D, NavVis}.
Given a complete 3D model, one can then, in principle,
predict the RF propagation between any two 
points in the environment with standard
ray tracing tools or other electro-magnetic (EM) solvers
\cite{yun2015ray}. 

\begin{figure}
    \centering
    \includegraphics[width=0.9\linewidth]{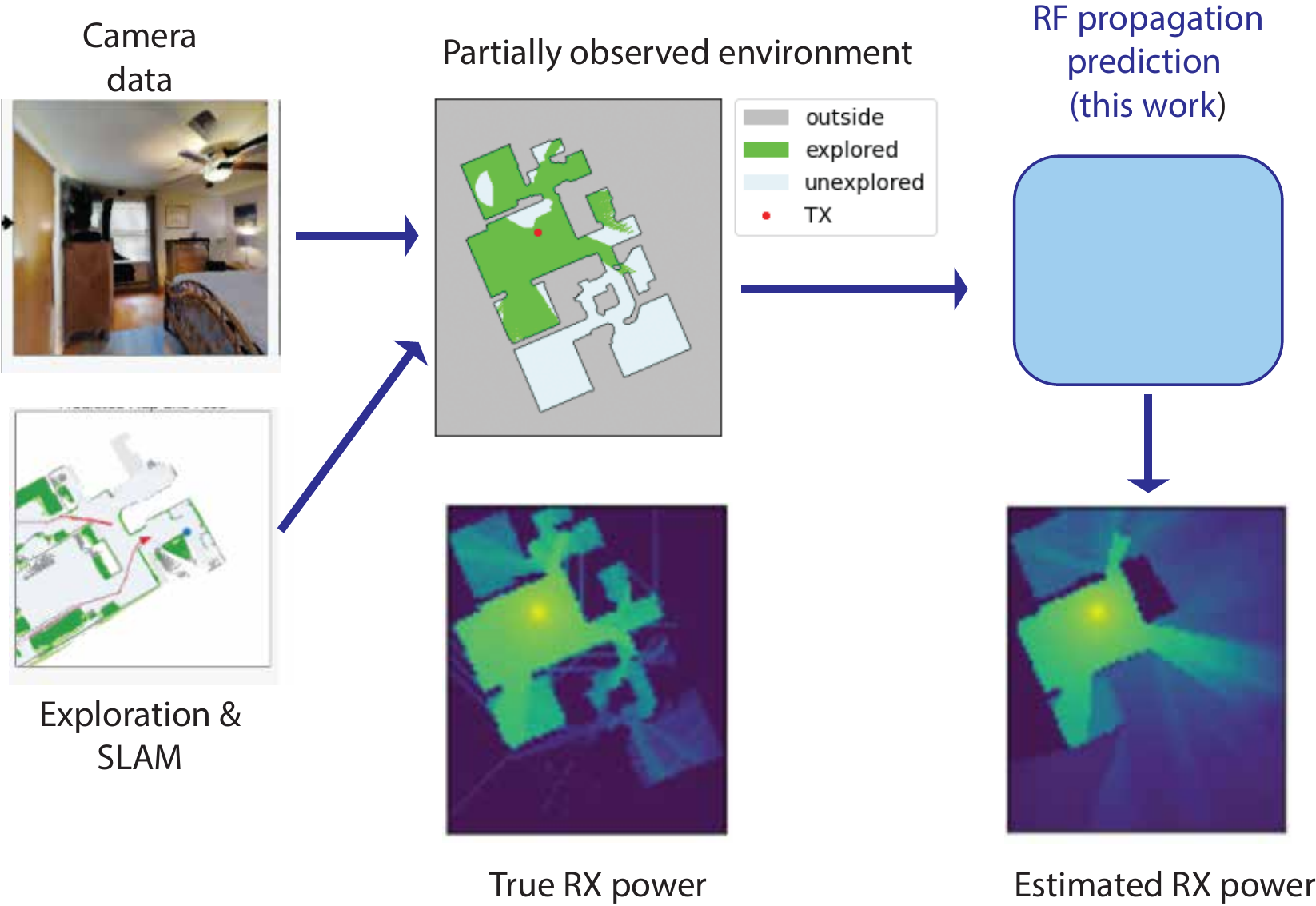}
    \caption{Example application: A partial map of an unknown
    environment is learned from camera data and robot navigation
    using tools such as Active Neural SLAM \cite{chaplot2020learning}.
    The problem is to estimate the RF propagation from a 
    hypothetical transmitter (TX) within the environment.
    The propagation should be estimated both in the 
    observed and unobserved areas.}
    \label{fig:overview}
    \vspace{-3mm}
\end{figure}

In this paper, we address a critical problem
that may arise in these applications:
\begin{quote}
\emph{
How do we predict RF propagation when
only a portion of the environment has been observed?}
\end{quote}
An example scenario for this problem is 
shown in Fig.~\ref{fig:overview}.
A robot agent explores an unknown indoor environment,
as may occur in search and rescue operations. Using
camera data, it builds a map of the environment
through some form of photogrammetry. The particular 
camera data
and partial map shown in Fig.~\ref{fig:overview} are sample output of a state-of-the-art Active Neural simultaneous localization and mapping (SLAM) 
module \cite{chaplot2020learning} simulated 
on the Gibson AI dataset \cite{xia2018gibson} that we will discuss
in detail below.  The agent is also given a hypothetical 
wireless transmitter (TX) location.  If the full environment
were known, the agent could determine the true receive (RX) power
throughout the environment. For example, the ``True RX power"
in Fig.~\ref{fig:overview} was predicted using a commercial ray tracing
solver \cite{Remcom}, also described in detail below.
The problem we address is how to estimate the propagation
when only a portion of the map has been reconstructed. Importantly,
we wish to estimate the propagation for links where the transmitter (TX) and/or
receiver (RX) are outside the environment that has been observed.

Finding good solutions to this problem is tremendously challenging,
particularly in the mmWave frequencies range,
which is the focus of this work.
In addition to the standard challenges of mmWave ray tracing 
\cite{lecci2021accuracy},
propagation prediction with partial information implicitly requires 
some model to ``fill in" the environment in the regions
that have not yet been observed. The problem can be seen 
as an EM ``inpainting", in analogy to the computer vision problem
where most state-of-the-art algorithms require 
extensively trained deep generative models
\cite{yu2018generative}.

As a starting point to understand the problem,
in this work, we consider a simple heuristic solution:
We first run ray tracing on the partial map, where the unexplored
area is treated as free space. We then train
simple machine learning-based estimators to predict link state (LOS, NLOS, or outage) and path gain
from features obtained by partial ray tracing.  

We show that the proposed methodology has a useful ``interpolation
property": On the one hand, when the environment is completely 
unobserved, the features are sufficient so that the predictor
can default to the form of a standard statistical channel model,
such as those used by 3GPP \cite{3GPP38901}.
On the other hand, as the partial information is built up,
the model can increasingly use the partial ray trace
to obtain more accurate, site-specific predictions.


\section{Problem Formulation}

\subsection{Environments and Partially Observed Environments}
We describe the \emph{environment} as 
some function of space, $F(\nbx)$, where 
$\nbx \in \R^d$ denotes the spatial position with $d=2$
or $d=3$ depending on whether we consider 2D or 3D
representations.  
For the purpose
of predicting EM propagation,
at each $\nbx$,
$F(\nbx)$ could be a binary variable indicating whether or not
a location is occupied or a discrete variable
with one of a finite set of values if there 
are multiple types of material. In addition,
$F(\nbx)$ may indicate the permittivity of
any surfaces and objects in the environment.

Given such an environment function $F(\nbx)$, 
we assume that
one can predict the true wireless channel between any $\nbx^t$ to $\nbx^r$.
Additionally, we assume that the channel is described by a standard multi-path 
ray model \cite{heath2018foundations} with parameters:
\begin{equation} \label{eq:path_param}
    \theta := \left\{ (a_\ell, \gamma_\ell), \ell = 1,\ldots,L \right\}
\end{equation}
where $L$ is the number of paths,
for each $\ell$, $a_\ell$ is the complex path gain, and $\gamma_\ell \subset \R^d$ is the route of the path
in space. We will assume that each route, $\gamma_\ell$,
consists of one or more line segments starting at the TX location,
$\nbx^t$, and ending at the RX location, $\nbx^r$. A path is line-of-sight (LOS)
if it consists of a single segment, that is, the path goes
directly from the TX to the RX. Otherwise, it is an non-LOS (NLOS) path.
From the path route, one can obtain the angles of departure,
angles of arrival, and delay, from which the wideband MIMO channel response for arbitrary arrays can be computed \cite{heath2018foundations}.


To model that fact that an environment is partially
observed, we assume there is a known set $A \subset \R^d$,
on which the environment has been observed. 
We call $A$ the \emph{observed area}. We also
assume that we have (a possibly approximate)
reconstruction of the environment in $A$ through some function,
$\widehat{F}(\nbx)$ for $\nbx \in A$.
 
\subsection{Link State and Omni-Directional Path Gain}

The general problem is to estimate some function
of the path parameters \eqref{eq:path_param}
 from the partially observed environment.  
We will denote this function by:
\begin{equation} \label{eq:path_fun}
    z = G( \theta ),
\end{equation}
and we call $z$, a \emph{channel statistic}. 
We model the problem probabilistically and treat $z$ as a random
variable with the goal of estimating
a conditional probability distribution:
\begin{equation} \label{eq:pzcond}
    p(z | A, \widehat{F} ),
\end{equation}
which represents the conditional distribution of the channel statistic
$z$, given the observed area, $A$, and the estimated environment $\widehat{F}(\nbx)$.
The randomness arises from the environment in the unobserved area
and noise or errors in the observed area. 

Although our methodology is general, 
in this work, we focus on estimating the channel statistic: 
\begin{equation} \label{eq:zsg}
    z = (s, g_{\rm omni}),
\end{equation}
where $s$ is the \emph{link state} and $g_{\rm omni}$ is the
\emph{omni-directional} path gain. For any channel between a 
TX and RX pair, we define its link state similar to \cite{akdeniz2014millimeter}.
Therefore, we will denote the link state by $s \in \{\mathrm{LOS},\mathrm{NLOS}, \mathrm{Outage}\}$.
Also, given a set of path parameters \eqref{eq:path_param},
we define the (clipped) omni-directional path gain as:
\begin{equation} \label{eq:omni_gain}
    g_{\rm omni} := \max\left\{ 10 \log_{10}\left[ \sum_{\ell=1}^L |a_\ell|^2 \right], g_{\rm min} \right\},
\end{equation}
which represents the average wideband 
path gain in dB seen when isotropic antennas are placed at the TX and RX.
To avoid singularities when the path gain is very low or when there are no paths,
we clip the path gain at a minimum value $g_{\rm min}$. For the remainder of the paper
we will use $g_{\rm min} = $\, \SI{-150}{dB}. Path gain values below this level have little
value for communication. We will use the convention that when a link is outage, that is, there are no paths, the omni-directional path gain will also be set to $g_{\rm min}$.

\section{Proposed Algorithm}

\subsection{Overview}
If the environment were completely known (that is, we knew
$F(\nbx)$ for all $\nbx$), in principle, one can find exactly
the channel parameters $\theta$ in \eqref{eq:path_param}
between any given TX and RX locations
by ray tracing or any other form of EM solver.  
Then, given the channel parameters, $\theta$,
theoretically one could exactly determine
any statistic, $z=G(\theta)$.

However, in the case of partial information on the environment,
estimation of the distribution of the statistic, $z$, is, in general, difficult.
Suppose that we have an observed area $A$ and an estimate of the environment, $\Fhat(\nbx)$ for $\nbx \in A$, and we wish
to compute the probability distribution $p(z|A,\Fhat)$.
In principle, one would need
a probabilistic model, $p(F|A,\Fhat)$, describing the distribution
of the true environment from the partially observed environment.
From such a distribution, one could theoretically sample $F$ and
compute the true channel parameters $\theta$ from the environment,
and the channel statistic $z=G(\theta)$ from the channel
parameters. However, obtaining a probabilistic model, $p(F|A,\Fhat)$,
is equivalent to rebuilding the true environment 
of the complete environment, $F$, from the partial
environment. Learning such a model would require tremendous
amounts of training data. Moreover, computing the
channel parameters for each potential true environment
would be computationally infeasible.

For the link state and omni-directional path gain,
$z=(s,g_{\rm omni})$,
we thus propose a simple heuristic algorithm.
We first run an approximate ray tracing on the observed 
component of the environment, treating the unobserved area
as free space. Then, we extract the key features of the 
estimated channel parameters from this partial map. Using these features
we train simple machine learning predictors for the channel
statistics.
We now describe each of these steps.



\subsection{Partial Map Path Prediction}

For the first phase, we extend $\widehat{F}(\nbx)$ to all of $\R^d$ by simply assuming that $F(\nbx)$
corresponds to free space for all $\nbx \not\in A$. Then, given the TX and RX locations, $\nbx^t$ and $\nbx^r$,
we run ray tracing, or any other EM simulation,
on this partial environment filled with free space to obtain an
initial set of path estimates. This simulation will obtain
a set of path parameters:
\begin{equation} \label{eq:path_fs}
    \theta_{\rm FS} := \left\{ (\widehat{a}_\ell, \widehat{\gamma}_\ell), \ell = 1,\ldots,\widehat{L} \right\}
\end{equation}
where $\widehat{L}$ is the number of paths in the free space-filled
environment, and, for each $\ell$, $\widehat{a}_\ell$ is the complex
path gain and $\widehat{\gamma}_\ell \subset \R^d$ is the path route 
in the space. We call $\theta_{\rm FS}$ the \emph{partial map path estimates}.

\subsection{Feature Extraction}

Our goal is to estimate the conditional probability $P(s,g_{\rm omni}|A,\Fhat)$ of the
link state $s$ and omni-directional path gain $g_{\rm omni}$ from the observed partial environment and $\theta_{\rm FS}$.
To simplify the prediction, we make the approximation that:
\begin{equation} \label{eq:Pcond_features}
    P(s, g_{\rm omni} | A, \Fhat) \approx P(s, g_{\rm omni} | \phi),
\end{equation}
where $\phi$ is a set of features extracted from the observed area $A$ and the estimated
environment $\Fhat(\nbx)$ on the observed area, $\nbx \in A$.  
In this work, we will explore a simple set of features:
\begin{equation}  \label{eq:features}
    \phi = (\shat, d_{\rm unobs}, d, \ghat_{\rm omni})
\end{equation}
where $\shat$ and $\ghat_{\rm omni}$ represent the estimated link state and omni-directional path gain in the partial environment respectively, $d$ is the distance between TX and RX, and $d_{\rm unobs}$ is the total unobserved distance along the LOS between TX and RX.
    

Although the features are simple, they are sufficient to provide good models for two extreme cases: when the environment is completely unobserved and when the environment is fully observed.
In the first case, we would expect to use a fully statistical model
such as the 3GPP models in \cite{3GPP38901}. In these models, the LOS probability and the
omni-directional path gain are both simply functions of the TX-RX distance $d$. 
In the other case, when the environment
is fully observed, we know that:
\begin{equation}
    (s,g_{\rm omni}) = (\shat, \ghat_{\rm omni})
\end{equation}
Hence, the features \eqref{eq:features} are also sufficient to provide accurate predictions when
the environment is fully observed. Finally, the characteristic $d_{\rm unobs}$ can be seen as a variable that represents the degree to which the relevant component of the environment is observed. A sufficiently rich predictor based on the above features can then interpolate 
between the fully observed and fully unobserved cases.

Now, we write the conditional distribution of $(s,g_{\rm omni})$ given the features $\phi$ as:
\begin{equation} \label{eq:psg_two}
    P(s,g_{\rm omni}|\phi) = P(s|\phi)P(g_{\rm omni}|s, \phi)
\end{equation}
We call the model that estimates the conditional probability $P(s|\phi)$ the \emph{link state classifier} and the model that estimates the conditional distribution 
$p(g_{\rm omni}|s,\phi)$ the \emph{omni-directional path gain predictor}. Both are described in the following subsections.

\subsection{Link State Classifier}

As described above, the link state classifier predicts the conditional probability:
\begin{equation} \label{eq:ls_prob}
    P(s|\phi)
\end{equation}
For this model, we obtained the best performance when considering only a subset of the features of \eqref{eq:features}, namely: LOS distance $d$, unobserved LOS distance $d_{\rm unobs}$, and free space-filled link state $\shat$. Regarding the structure of the model, for each value of $\shat$, we build a simple logistic classifier based on the two features $d_{\rm unobs}$ and $d$.

\begin{table}[t]
    \centering
    \caption{Parameters of omni-directional path gain predictors}
    \label{tab:path_loss_network}
    
    \begin{tabular}{|>{\raggedright}m{1.3cm}|>{\raggedright}m{1.3cm}|>{\raggedright}m{1.4cm}|>{\raggedright}m{1cm}|>{\raggedright}m{1cm}|}
    \hline
    Estimated link state $s$ & Partial link state $\shat$ & Input features & Hidden layers & Hidden units
    \tabularnewline \hline
    LOS & LOS & $d$, $d_{\rm unobs}$, $\ghat_{\rm omni}$& 2 & 20
        \tabularnewline \hline
    LOS & NLOS, Out & $d$  & 2 & 10         
        \tabularnewline \hline
    NLOS & NLOS, Out & $d$, $d_{\rm unobs}$, $\ghat_{\rm omni}$ & 2 & 20
        \tabularnewline \hline
    NLOS & LOS & $d$  & 2 & 10 
        \tabularnewline \hline
    Out & Any & None & None & None
        \tabularnewline \hline
    \end{tabular}
    
\end{table}

\subsection{Omni-directional Gain Predictor}

Ideally, the omni-directional gain predictor should estimate the conditional distribution
$p(g_{\rm omni}|\phi,s)$, representing the conditional distribution of the omni-directional path gain
given the features $\phi$ and estimated link state $s$, given by the link state classifier.  
To simplify the model, we assume that it is Gaussian distributed. In this way, we can simply predict its mean and logarithmic variance:
\[
    \mathbb{E}( g_{\rm omni}|s,\phi), \quad
    \log \mathrm{var}(g_{\rm omni}|s,\phi).
\]
When the estimated link state is outage ($s=$ Outage), by convention described above, the path gain is set to the minimum value $g_{\rm min}$. For each of the other four cases as reported in Table~\ref{tab:path_loss_network}, we train a simple neural network depending on the estimated link state $s$ and partial link state $\shat$. When the partial link state matches the estimated link state, we use the partial map features such as $d_{\rm unobs}$ and $\ghat_{\rm omni}$. However, when they do not match, we simply use $d$, ignoring any non-reliable partial map information.

\iftoggle{conf}{}{

\subsection{Indoor-Outdoor Estimation}
In the training data set, all the TX-RX pairs are
both indoor.  However, when an environment is unexplored,
the boundaries of the building are not known, so a given
TX or RX location may be outside. 
We briefly discuss how to modify the classifier
when the TX is known to be inside, but the RX is potentially
outside.

Assume the TX is inside and 
let $S$ be the event that the RX location is inside.
The previous method trains a classifier 
to predict the conditional
probability 
\begin{equation} \label{eq:Pscond}
    P(s | \shat, d_{\rm unobs}, S),
\end{equation}
which is the conditional distribution on the true link state,
given the link state on the partial map, and the total
unobserved distance, when the RX is known to be inside.  
To compute the conditional probability
when the RX may be outside, we use the total probability,
\begin{align} \label{eq:Pstot}
    \MoveEqLeft P(s | \shat, d_{\rm unobs}) = P(s | \shat, d_{\rm unobs},S)
    P(S| \shat, d_{\rm unobs}) \nonumber \\
    &+ P(s | \shat, d_{\rm unobs},S^c)
    P(S^c| \shat, d_{\rm unobs}).
\end{align}
We then estimate the conditional probability
\begin{equation} \label{eq:PScond}
    P(S|\shat, d_{\rm unobs})
\end{equation} using a \label{eq:PScond}
Simple logistic classifiers trained on links where
the RX locations are uniformly selected in the map area
and hence have cases where the RX is inside and outside.
This classification does not require running any
ray tracing since all quantities are geometric.

Furthermore, since we focus on the mmWave frequency range,
the indoor-outdoor penetration has minimal \cite{Rappaport2014-mmwbook}. Therefore, as a simplification,
we will assume the outdoor RX locations are in outage.
Therefore, 
\begin{equation} \label{eq:Pouts}
    P(s=\mathrm{Out} | \shat, d_{\rm unobs}, S^c) = 1
\end{equation}
fro all values of $\shat$ and $d_{\rm unobs}$.   
Applying \eqref{eq:Pouts} along with the learned distributions
\eqref{eq:Pscond} and \label{eq:PScond},
we can then compute \eqref{eq:Pstot}.

}

\section{Results}
\begin{figure}
  \centering
  \includegraphics[width=0.9
  \linewidth]{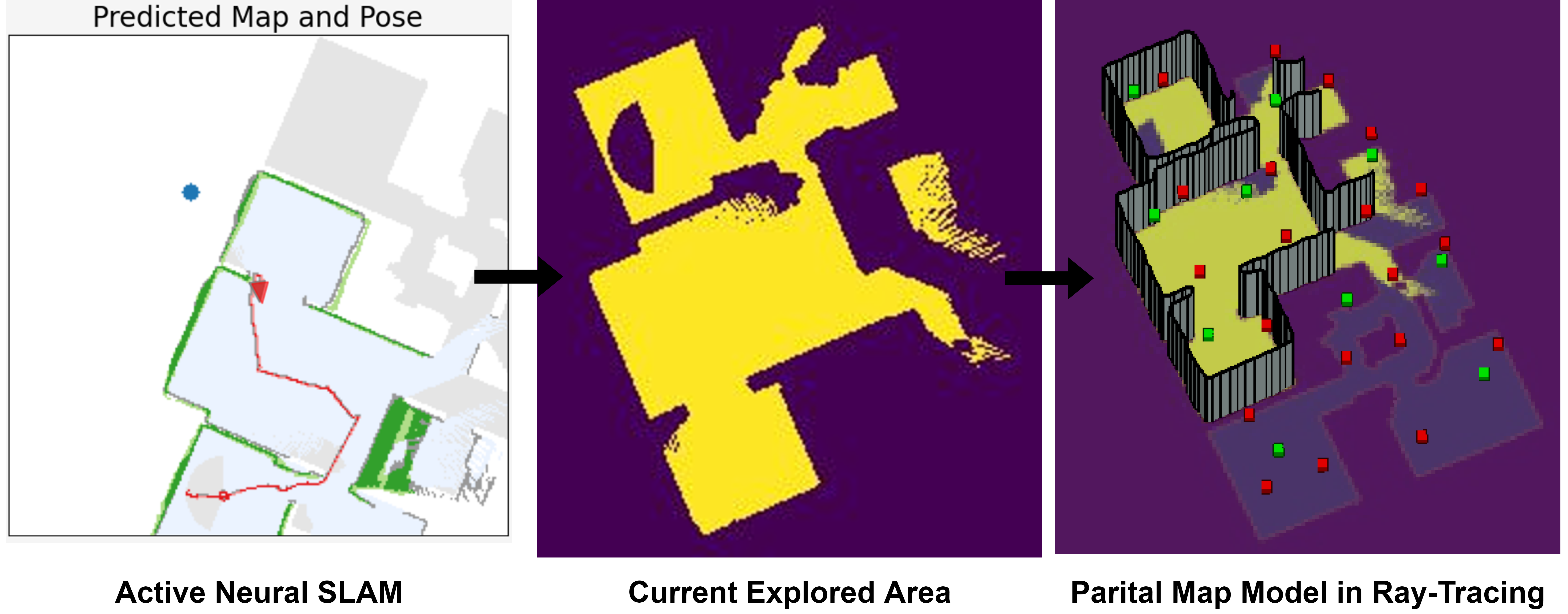}
  \caption{We constructed partial maps by exporting the explored regions of an indoor environment from the Active Neural SLAM procedure. We then fed this map into a ray  tracing tool.}
  \vspace{-3mm}
  \label{fig:partial_map}
\end{figure}
\subsection{Dataset and Training}

The methods are validated on a realistic
simulation of robot navigation and visual
SLAM map construction, used in prior 
work \cite{yin2022millimeter}.
We used
the Gibson indoor dataset which contains 
a collection of accurate 3D maps 
along with camera data \cite{xia2018gibson}.
We used 8 of the environments for training
and 4 for testing. 
Within each training environment, we selected
400 random TX-RX location pairs, and within each
test environment.
We performed ray tracing simulations using  Remcom Wireless InSite \cite{Remcom} at \SI{28}{GHz} to obtain ground truth values for
the channels in each TX-RX link. Then, 
similar to \cite{yin2022millimeter}, we use the 
Active Neural SLAM algorithm \cite{chaplot2020learning} 
to simulate robot indoor exploration on the AI Habitat platform from visual information~\cite{habitat19iccv}. The 
robot explores the environment and gradually builds a simplified 3D map.  
As shown in Fig.~\ref{fig:partial_map}, we constructed partial maps by exporting the explored region of an environment from the Active Neural SLAM procedure.  
We extracted partial maps from four stages: 50, 100, 150, and 200 steps, where 200 steps generally correspond to approximately 60\% of the total
indoor area.
The higher the step count, the closer we get to observing a complete map. The partial maps along
with the TX and RX locations served as input to the channel prediction algorithms that were trained
to estimate the channel. Links were created
where the TX and RX were both inside and outside
the observed area, and in LOS, NLOS, and outage conditions.
  

\begin{figure}
    \centering
    \includegraphics[width=0.6\linewidth]{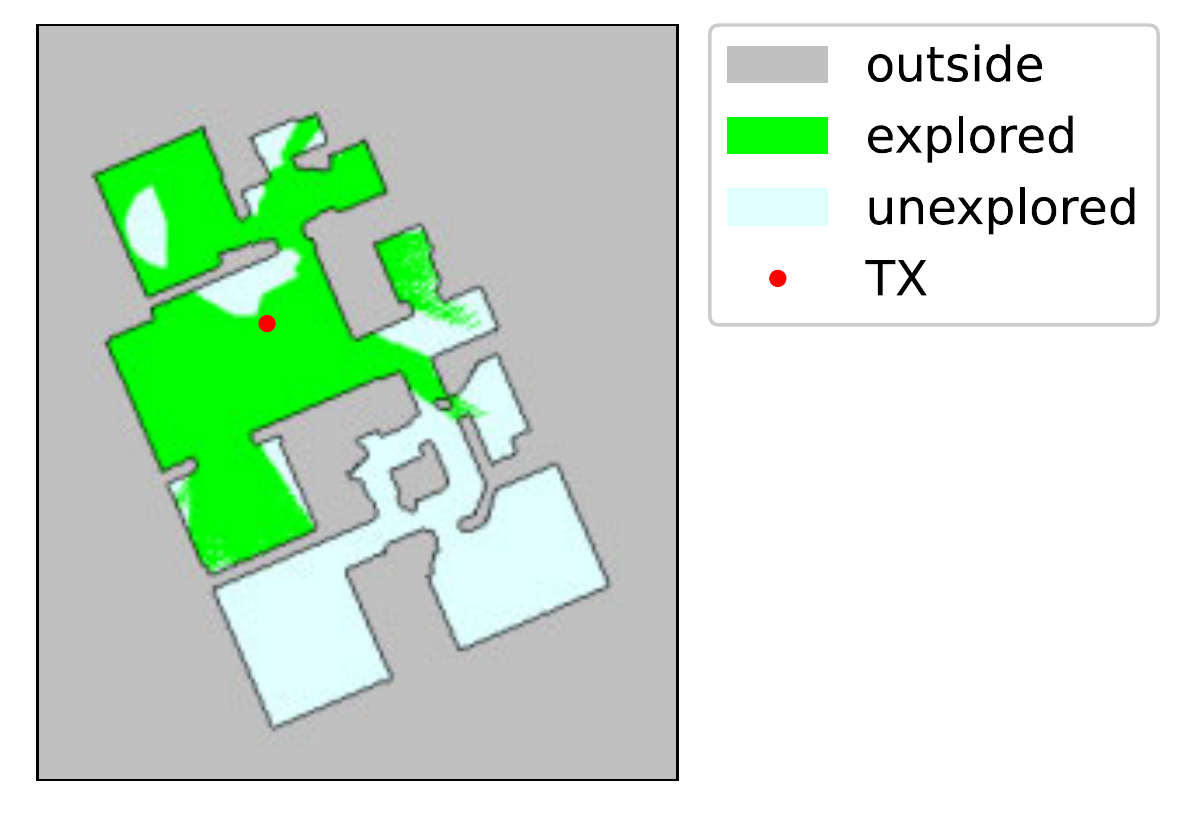}
    \caption{Example of test area from the \emph{Shelbiana} map
    in the Gibson dataset \cite{xia2018gibson}. This
    map was not in the training data. The displayed area
    is $11.4 \times 13.4$\,\si{\meter\squared}.
    The TX location is shown in red.}
    \label{fig:example_map}
\end{figure}



\begin{figure} 
    \centering
  \subfloat[]{%
       \includegraphics[width=\linewidth]{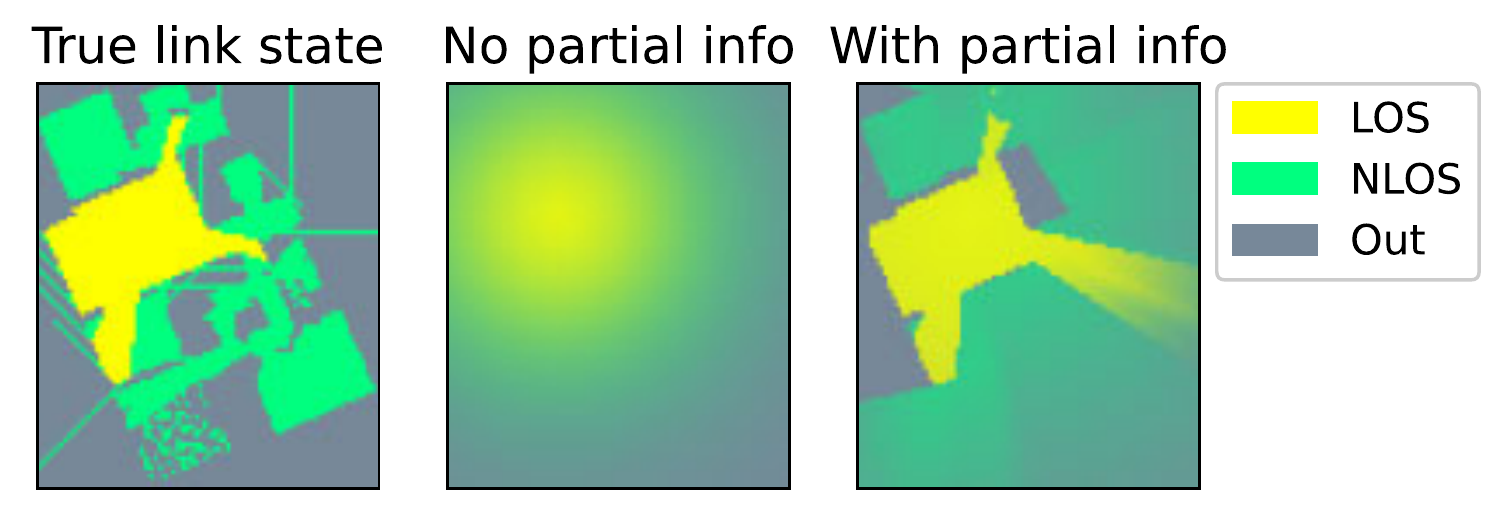}}
    \\ \vspace{-4mm}
  \subfloat[]{%
        \includegraphics[width=\linewidth]{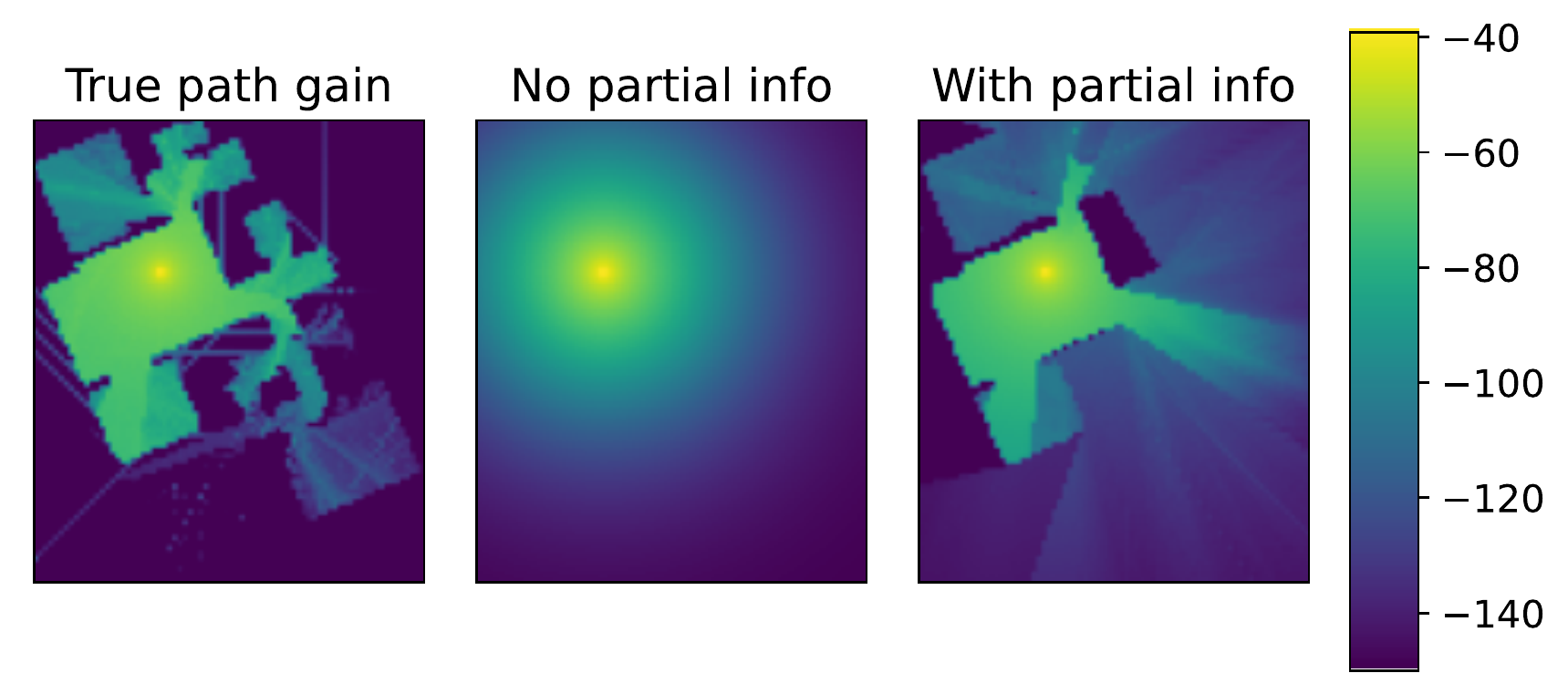}}

  \caption{(a) True link state (left) for the 
    test map in Fig.~\ref{fig:example_map} along with predicted
    link state using no partial information (center) and the partial information (right).
    (b) True omni-directional path gain (left) for the 
    test map in Fig.~\ref{fig:example_map} along with predicted
    path gain using no partial information (center) and the partial information (right).}
    \vspace{-3mm}
  \label{fig:map_ls_pl} 
\end{figure}

\subsection{Example Prediction}

To illustrate the trained propagation predictor, we first
consider an example of test environment shown in Fig.~\ref{fig:example_map}.
This area is one of the maps in the Gibson dataset that 
was \emph{not} used during training and is therefore representative of the predictor's ability to generalize to completely new environments.
In this case, the Neural SLAM ran over 150 steps, which led to an observation of 51\% of the interior space. 

We ran ray tracing on the full one-layer simplified \emph{Shelbiana} environment from the TX location in Fig.~\ref{fig:example_map} 
to 6840 RX locations that are deployed in a grid with \SI{15}{\cm} spacing.

The left panel of Fig.~\ref{fig:map_ls_pl}a shows the true link state (LOS, NLOS, or outage) at each RX location.  
The middle panel shows the predicted 
link state of the trained model 
when there is no partial information. Since the model outputs
a probability, the color shown is the sum of the colors
for the three link states, weighted by their probabilities.
With no information in the environment, the predicted link state
is only a function of the distance from the TX, since there
is no environmental information to use.

The right panel of Fig.~\ref{fig:map_ls_pl}a shows the 
predicted link state with the partial information. We see
that the link state is well predicted in the observed area, as expected. As the RX locations move out of the observed area,
 uncertainty increases and 
the prediction becomes less accurate.

Similarly, Fig.~\ref{fig:map_ls_pl}b depicts the true and predicted
omni-directional path gain. A similar pattern can be observed
with the link state comparing the true omni-directional
path gain with the predictions with and without partial information.

\begin{figure}
    \centering
    \includegraphics[width=0.85\linewidth]{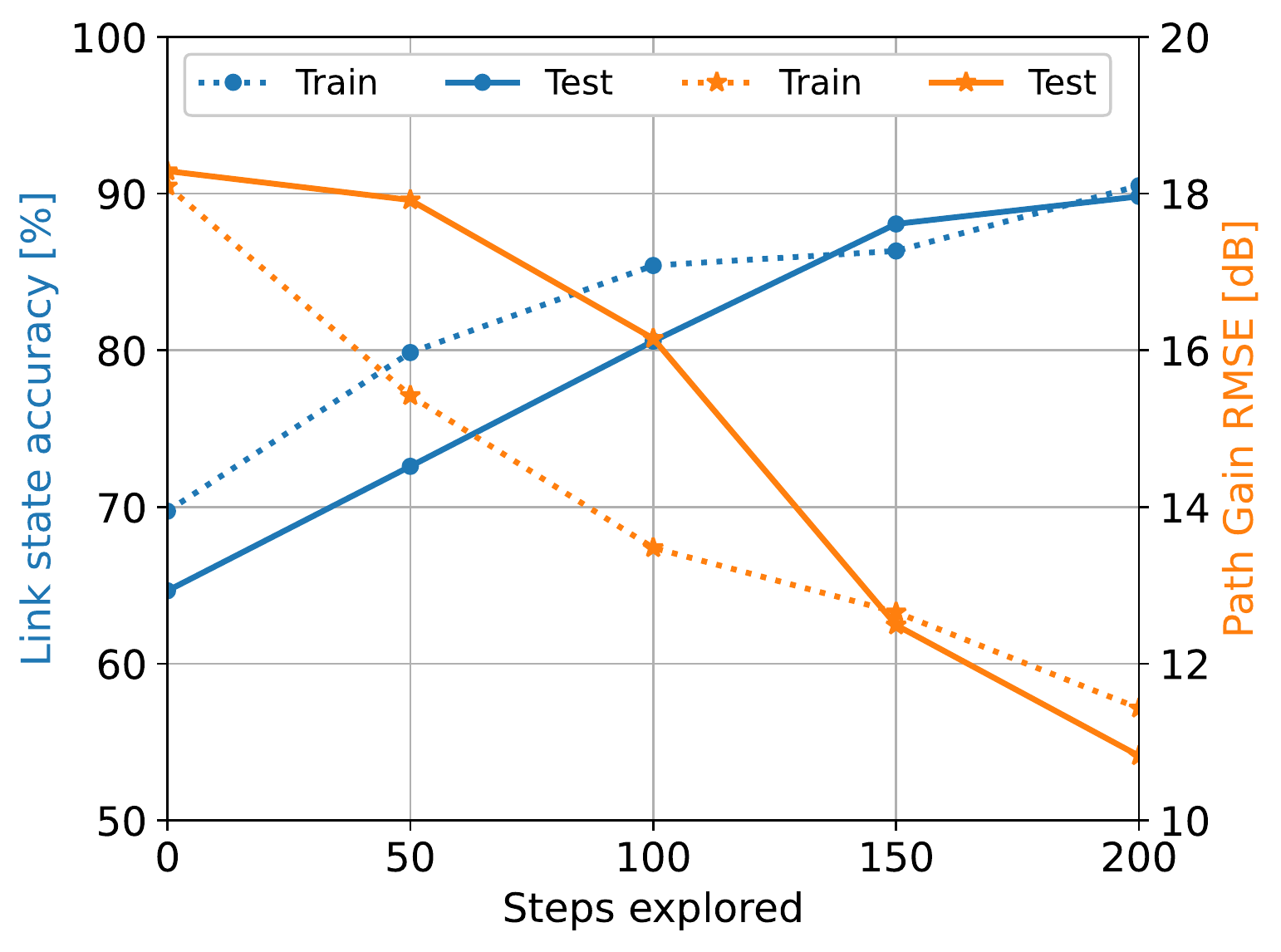}
    \caption{Blue: Accuracy of the link 
    state prediction. Orange:  
    Root mean squared error (RMSE) of the path gain. 
    Both metrics are plotted as a function of
    the number of exploration steps in the environment.}
    \vspace{-4mm}
    \label{fig:perf_step}
\end{figure}

\subsection{Average Test Performance}

To quantify the average performance,
Fig.~\ref{fig:perf_step} shows the accuracy of link state estimation and the root mean square error of the omni-directional path gain as a function of the number
of steps the robot has explored in the environment. The two metrics are plotted as the average performance on both the 8 training maps and 4 test maps.
The difference in performance between 
the training and test data is due to the considerable variations between different interior spaces.

When step $=0$, the environment is completely unexplored and the link state and the path gain must be estimated from the TX-RX distance only. 
This type of model is similar to the case of fully statistical 
3GPP models \cite{3GPP38901} that do not account for the 3D geometry of the environment.  
In this case, we obtain a test RMSE of the path gain of 
approximately \SI{18}{dB} and test link state accuracy of only 
approximately 65\%.
As the robot explores the environment (steps $=200$), the test RMSE for the path gain reduces to approximately \SI{11}{dB} whereas the link state accuracy peaks at approximately 90\%.

\section{Conclusions}

We have formulated a novel problem of 
estimating RF propagation in partially observed environments.
Solutions to this problem may be valuable in scenarios where
mobile agents have access to camera data that can assist RF
communication and sensing, but the environment is not fully 
observed. We proposed a simple solution that improves channel prediction and allows us to assess the degree of uncertainty
to guide further exploration.  
Future work will investigate more sophisticated 
models. For example, the explored area can be
converted into image tensors so that radio propagation can be predicted as an image regression problem using deep convolutional neural networks, as proposed in \cite{zhang2020cellular}.

\bibliographystyle{IEEEtran}
\bibliography{bibl}

\end{document}

%% file: commands.tex
\def\nbx{{\mathbf{x}}}

\def\nb0{{\mathbf{0}}}
\def\nb1{{\mathbf{1}}}

\def\R{\mathbb{R}}

\def\R{{\mathbb{R}}}

\def\ghat{\widehat{g}}
\def\shat{\widehat{s}}

\def\Fhat{\widehat{F}}

%% file: acronyms.tex
\usepackage[acronyms,nonumberlist,nopostdot,nomain,nogroupskip]{glossaries}
\newacronym{quic}{QUIC}{Quick UDP Internet Connections}
\newacronym{3gpp}{3GPP}{3rd Generation Partnership Project}
\newacronym{adc}{ADC}{Analog to Digital Converter}
\newacronym{5g}{5G}{5th generation}
\newacronym{aimd}{AIMD}{Additive Increase Multiplicative Decrease}
\newacronym{am}{AM}{Acknowledged Mode}
\newacronym{amc}{AMC}{Adaptive Modulation and Coding}
\newacronym{aqm}{AQM}{Active Queue Management}
\newacronym{awgn}{AGWN}{Additive White Gaussian Noise}
\newacronym{afd}{AFD}{Austin Fire Department}
\newacronym{balia}{BALIA}{Balanced Link Adaptation}
\newacronym{bdp}{BDP}{Bandwidth-Delay Product}
\newacronym{bf}{BF}{Beamforming}
\newacronym{cc}{CC}{Congestion Control}
\newacronym{cdf}{CDF}{Cumulative Distribution Function}
\newacronym{cn}{CN}{Core Network}
\newacronym{cqi}{CQI}{Channel Quality Information}
\newacronym{cp}{CP}{Control Plane}
\newacronym{csirs}{CSI-RS}{Channel State Information - Reference Signal}
\newacronym{dc}{DC}{Dual Connectivity}
\newacronym{dce}{DCE}{Direct Code Execution}
\newacronym{dci}{DCI}{Downlink Control Information}
\newacronym{dl}{DL}{Downlink}
\newacronym{dmr}{DMR}{Deadline Miss Ratio}
\newacronym{dmrs}{DMRS}{DeModulation Reference Signal}
\newacronym{e2e}{E2E}{End-to-End}
\newacronym{ecn}{ECN}{Explicit Congestion Notification}
\newacronym{edf}{EDF}{Earliest Deadline First}
\newacronym{enb}{eNB}{evolved Node Base}
\newacronym{epc}{EPC}{Evolved Packet Core}
\newacronym{es}{ES}{Edge Server}
\newacronym{fdma}{FDMA}{Frequency Division Multiple Access}
\newacronym{fdd}{FDD}{Frequency Division Duplexing}
\newacronym[firstplural=Radio Access Technologies (RATs)]{rat}{RAT}{Radio Access Technology}
\newacronym{fs}{FS}{Fast Switching}
\newacronym{ftp}{FTP}{File Transfer Protocol}
\newacronym{gnb}{gNB}{Next Generation Node Base}
\newacronym{harq}{HARQ}{Hybrid Automatic Repeat reQuest}
\newacronym{hetnet}{HetNet}{Heterogeneous Network}
\newacronym{hh}{HH}{Hard Handover}
\newacronym{hol}{HOL}{Head-of-Line}
\newacronym{ia}{IA}{Initial Access}
\newacronym{imt}{IMT}{International Mobile Telecommunication}
\newacronym{iot}{IoT}{Internet of Things}
\newacronym{los}{LOS}{Line of Sight}
\newacronym{lte}{LTE}{Long Term Evolution}
\newacronym{m2m}{M2M}{Machine to Machine}
\newacronym{mac}{MAC}{Medium Access Control}
\newacronym{mc}{MC}{Multi-Connectivity}
\newacronym{mcs}{MCS}{Modulation and Coding Scheme}
\newacronym{mec}{MEC}{Mobile Edge Cloud}
\newacronym{mi}{MI}{Mutual Information}
\newacronym{mimo}{MIMO}{Multiple Input, Multiple Output}
\newacronym{mmwave}{mmWave}{millimeter wave}
\newacronym{mr}{MR}{Maximum Rate}
\newacronym{mss}{MSS}{Maximum Segment Size}
\newacronym{mtd}{MTD}{Machine-Type Device}
\newacronym{mtu}{MTU}{Maximum Transmission Unit}
\newacronym{nfv}{NFV}{Network Function Virtualization}
\newacronym{nlos}{NLOS}{Non Line of Sight}
\newacronym{nr}{NR}{New Radio}
\newacronym{ofdm}{OFDM}{Orthogonal Frequency Division Multiplexing}
\newacronym{pdcch}{PDCCH}{Physical Downlonk Control Channel}
\newacronym{pdcp}{PDCP}{Packet Data Convergence Protocol}
\newacronym{pdsch}{PDSCH}{Physical Downlink Shared Channel}
\newacronym{pdu}{PDU}{Packet Data Unit}
\newacronym{pf}{PF}{Proportional Fair}
\newacronym{pgw}{PGW}{Packet Gateway}
\newacronym{phy}{PHY}{Physical}
\newacronym{pbch}{PBCH}{Physical Broadcast Channel}
\newacronym[plural=\gls{mme}s,firstplural=Mobility Management Entities (MMEs)]{mme}{MME}{Mobility Management Entity}
\newacronym{prb}{PRB}{Physical Resource Block}
\newacronym{pss}{PSS}{Primary Synchronization Signal}
\newacronym{pucch}{PUCCH}{Physical Uplink Control Channel}
\newacronym{pusch}{PUSCH}{Physical Uplink Shared Channel}
\newacronym{rach}{RACH}{Random Access Channel}
\newacronym{ran}{RAN}{Radio Access Network}
\newacronym{red}{RED}{Robotics Emergency Deployment}
\newacronym{rf}{RF}{Radio Frequency}
\newacronym{rlc}{RLC}{Radio Link Control}
\newacronym{rlf}{RLF}{Radio Link Failure}
\newacronym{rrc}{RRC}{Radio Resource Control}
\newacronym{rrm}{RRM}{Radio Resource Management}
\newacronym{rr}{RR}{Round Robin}
\newacronym{rs}{RS}{Remote Server}
\newacronym{rsrp}{RSRP}{Reference Signal Received Power}
\newacronym{rss}{RSS}{Received Signal Strength}
\newacronym{rtt}{RTT}{Round Trip Time}
\newacronym{rw}{RW}{Receive Window}
\newacronym{rx}{RX}{Receiver}
\newacronym{sa}{SA}{standalone}
\newacronym{sack}{SACK}{Selective Acknowledgment}
\newacronym{sap}{SAP}{Service Access Point}
\newacronym{sch}{SCH}{Secondary Cell Handover}
\newacronym{scoot}{SCOOT}{Split Cycle Offset Optimization Technique}
\newacronym{sdma}{SDMA}{Spatial Division Multiple Access}
\newacronym{sinr}{SINR}{Signal to Interference plus Noise Ratio}
\newacronym{sm}{SM}{Saturation Mode}
\newacronym{snr}{SNR}{Signal to Noise Ratio}
\newacronym{son}{SON}{Self-Organizing Network}
\newacronym{ss}{SS}{Synchronization Signal}
\newacronym{srs}{SRS}{Sounding Reference Signal}
\newacronym{sss}{SSS}{Secondary Synchronization Signal}
\newacronym{tb}{TB}{Transport Block}
\newacronym{tcp}{TCP}{Transmission Control Protocol}
\newacronym{tdd}{TDD}{Time Division Duplexing}
\newacronym{tdma}{TDMA}{Time Division Multiple Access}
\newacronym{tfl}{TfL}{Transport for London}
\newacronym{tm}{TM}{Transparent Mode}
\newacronym{trp}{TRP}{Transmitter Receiver Pair}
\newacronym{tti}{TTI}{Transmission Time Interval}
\newacronym{ttt}{TTT}{Time-to-Trigger}
\newacronym{tx}{TX}{Transmitter}
\newacronym{ue}{UE}{User Equipment}
\newacronym{ul}{UL}{Uplink}
\newacronym{uml}{UML}{Unified Modeling Language}
\newacronym{um}{UM}{Unacknowledged Mode}
\newacronym{utc}{UTC}{Urban Traffic Control}
\newacronym{vm}{VM}{Virtual Machine}
\newacronym{rsrq}{RSRQ}{Reference Signal Received Quality}
\newacronym{rssi}{RSSI}{Received Signal Strength Indicator}
\newacronym{crs}{CRS}{Cell Reference Signal}
\newacronym{comp}{CoMP}{Coordinated Multi-Point}
\newacronym{cran}{C-RAN}{Cloud \acrlong{ran}}
\newacronym{ca}{CA}{Carrier Aggregation}
\newacronym{cco}{CC}{Carrier Component}
\newacronym{nsa}{NSA}{Non Stand Alone}
\newacronym{embb}{eMBB}{Enhanced Mobility Broadband}
\newacronym{bsr}{BSR}{Buffer Status Report}
\newacronym{srb}{SRB}{Service Radio Bearer}
\newacronym{scm}{SCM}{Spatial Channel Model}
\newacronym{sctp}{SCTP}{Stream Control Transmission Protocol}
\newacronym{mptcp}{MPTCP}{Multi-path TCP}
\newacronym{ietf}{IETF}{Internet Engineering Task Force}
\newacronym{os}{OS}{Operating System}
\newacronym{tls}{TLS}{Transport Layer Security}
\newacronym{rfc}{RFC}{Request for Comments}
\newacronym{http}{HTTP}{HyperText Transfer Protocol}
\newacronym{nat}{NAT}{Network Address Translation}
\newacronym{api}{API}{Application Programming Interface}
\newacronym{rto}{RTO}{Retransmission Timeout}
\newacronym{psc}{PSC}{Public Safety Communication}
\newacronym{rpgm}{RPGM}{Reference Point Group Mobility}
\newacronym{ic}{IC}{Incident Command}
\newacronym{rsu}{RSU}{Road Side Unit}
\newacronym{uav}{UAV}{unmanned aerial vehicle}
\newacronym{usv}{USV}{Unmanned Surface Vehicle}
\newacronym{uas}{UAS}{Unmanned Aerial System}
\newacronym{iab}{IAB}{Integrated Access and Backhaul}
\newacronym{qoe}{QoE}{Quality of Experience}
\newacronym{ssim}{SSIM}{Structural Similarity Index}
\newacronym{psnr}{PSNR}{Peak Signal to Noise Ratio}
\newacronym{bs}{BS}{Base Station}
\newacronym{mu}{MU}{Multiple User}
\newacronym{ag}{AG}{Air-to-Ground}
\newacronym{af}{AF}{Array Factor}
\newacronym{ula}{ULA}{Uniform Linear Array}
\newacronym{upa}{UPA}{Uniform Planar Array}
\newacronym{lcs}{LCS}{Local Coordinate System}
\newacronym{psd}{PSD}{Power Spectral Density}
\newacronym{vq}{VQ}{vector quantization}
\newacronym{a2g}{A2G}{air-to-ground}
\newacronym{em}{EM}{electromagnetic}
\newacronym{vae}{VAE}{variational autoencoder}